\documentclass[sigconf]{acmart}

\usepackage{colortbl}
\usepackage{multirow}
\AtBeginDocument{%
  }

\begin{document}

\title{Bridging the Micro--Macro Gap: Frequency-Aware Semantic Alignment for Image Manipulation Localization}

\author{Xiaojie Liang}
\orcid{0009-0005-5342-0003}
\email{liangxj53@mail2.sysu.edu.cn}
\affiliation{%
\institution{Sun Yat-sen University}
\department{School of Computer Science and Engineering}
\city{Guangzhou}
\state{Guangdong}
\country{China}
}

\author{Zhimin Chen}
\orcid{0009-0008-7350-8784}
\email{chenzhm56@mail2.sysu.edu.cn}
\affiliation{%
\institution{Sun Yat-sen University}
\department{School of Computer Science and Engineering}
\city{Guangzhou}
\state{Guangdong}
\country{China}
}

\author{Ziqi Sheng}
\correspondingauthor
\orcid{0009-0005-5106-9563}
% \authornote{Corresponding author.}
\email{shengzq3@mail.sysu.edu.cn}
\affiliation{%
\institution{Sun Yat-sen University}
\department{School of Computer Science and Engineering}
\city{Guangzhou}
\state{Guangdong}
\country{China}
}

\author{Wei Lu}
\orcid{0000-0002-4068-1766}
\email{luwei3@mail.sysu.edu.cn}
\affiliation{
\institution{Sun Yat-sen University}
\department{School of Computer Science and Engineering}
\city{Guangzhou}
\state{Guangdong}
\country{China}
}

\renewcommand{\shortauthors}
{Xiaojie Liang, Zhimin Chen, Ziqi Sheng, and Wei Lu}

\begin{abstract}
  As generative image editing advances, image manipulation localization (IML) must handle both traditional manipulations with conspicuous forensic artifacts and diffusion-generated edits that appear locally realistic. Existing methods typically rely on either low-level forensic cues or high-level semantics alone, leading to a fundamental micro--macro gap. To bridge this gap, we propose FASA, a unified framework for localizing both traditional and diffusion-generated manipulations. Specifically, we extract manipulation-sensitive frequency cues through an adaptive dual-band DCT module and learn manipulation-aware semantic priors via patch-level contrastive alignment on frozen CLIP representations. We then inject these priors into a hierarchical frequency pathway through a semantic-frequency side adapter for multi-scale feature interaction, and employ a prototype-guided, frequency-gated mask decoder to integrate semantic consistency with boundary-aware localization for tampered region prediction. Extensive experiments on OpenSDI and multiple traditional manipulation benchmarks demonstrate state-of-the-art localization performance, strong cross-generator and cross-dataset generalization, and robust performance under common image degradations.
\end{abstract}

\ccsdesc[500]{Computing methodologies~Artificial intelligence}
\ccsdesc[500]{Security and privacy~Intrusion/anomaly detection and malware mitigation}

\keywords{Image forgery detection, image manipulation localization}


\maketitle

\begin{figure}[t]
  \centering
  \includegraphics[width=\linewidth]{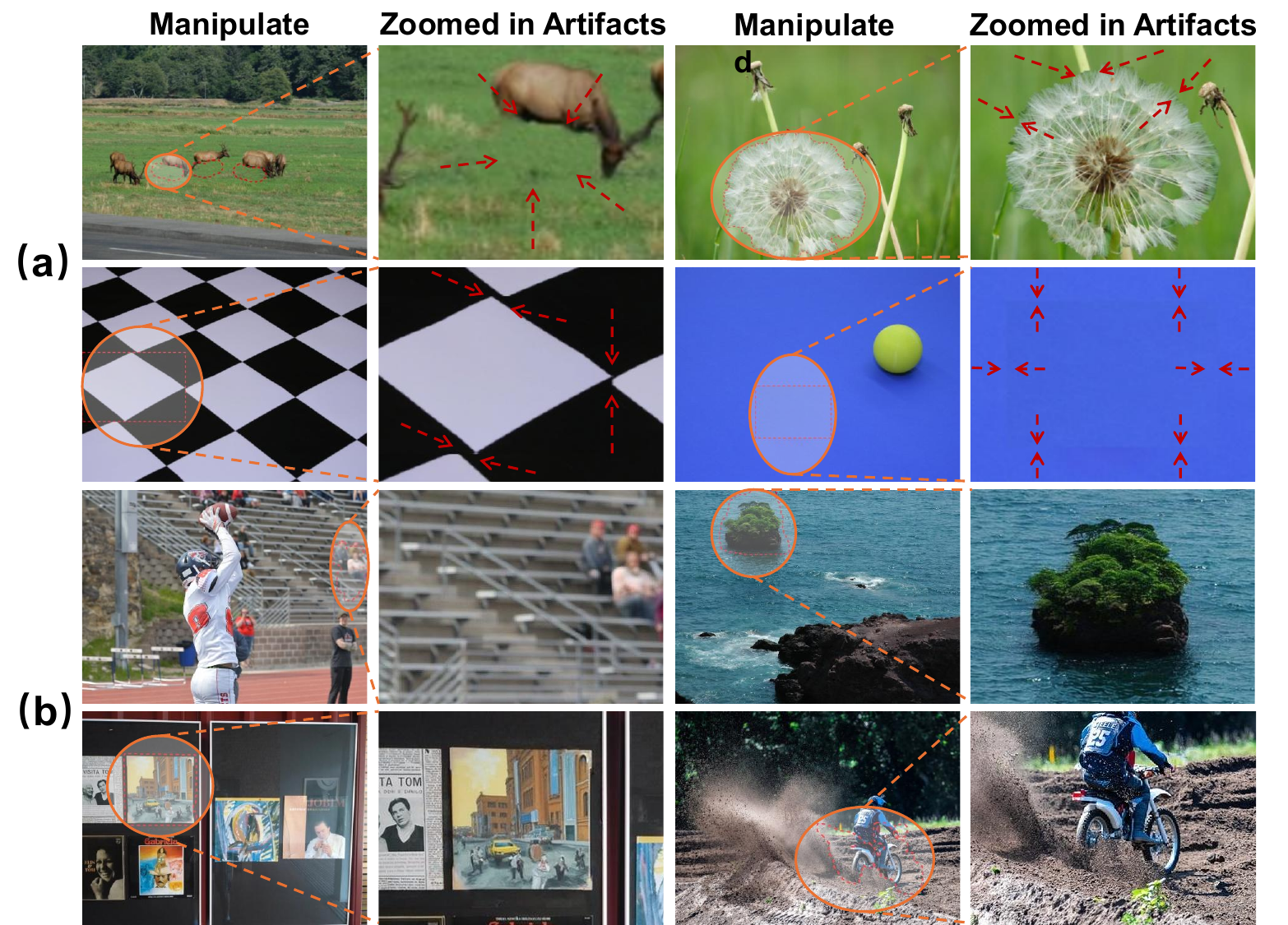}
  \Description{A side-by-side visual comparison of image manipulations. On the left, traditional manipulations show conspicuous microscopic artifacts such as unnatural edges and noise. On the right, diffusion-generated edits display locally realistic content that blends seamlessly with the background, showing coherent low-level statistics without obvious visible artifacts.}
  \caption{\textbf{Comparison of manipulation traces.} (a) Traditional manipulations leave conspicuous artifacts. (b) Diffusion-generated edits produce locally realistic content with coherent low-level statistics.}
  \label{fig:visual_gap}
\end{figure}

\section{Introduction}
With the rapid advancement of generative models and image editing tools, manipulated images pose growing threats to media credibility and information security~\cite{mehrjardi2023survey}. Image manipulation localization (IML) identifies tampered regions at the pixel level and provides fine-grained forensic evidence. Related forensic tasks have also received increasing attention~\cite{qu2023towards,qu2025revisiting,qu2026textshield}.

Most existing IML methods~\cite{wu2019mantra, li2022image, wu2022robust, shi2023pl, zanardelli2023image, zhu2026progressive, sheng2024dirloc, zeng2024mgqformer, liu2024multi, zhang2025edge, qu2026omni} target traditional manipulations (e.g., splicing, copy-move, inpainting~\cite{kaur2023image}) that disrupt intrinsic image statistics and leave microscopic forensic traces like abnormal noise patterns and frequency artifacts, achieving strong performance on traditional benchmarks. Conversely, diffusion-generated edits introduce a substantially different challenge. As shown in Figure~\ref{fig:visual_gap}, diffusion models alter high-level semantics while generating locally realistic content with coherent low-level statistics. This shift reduces the effectiveness of traditional IML methods, necessitating models that jointly exploit low-level forensic evidence and high-level semantic cues. The challenge has recently become more prominent with the introduction of benchmarks such as OpenSDI~\cite{wang2025opensdi}.

\begin{figure}[t]
  \centering
  \includegraphics[width=\linewidth]{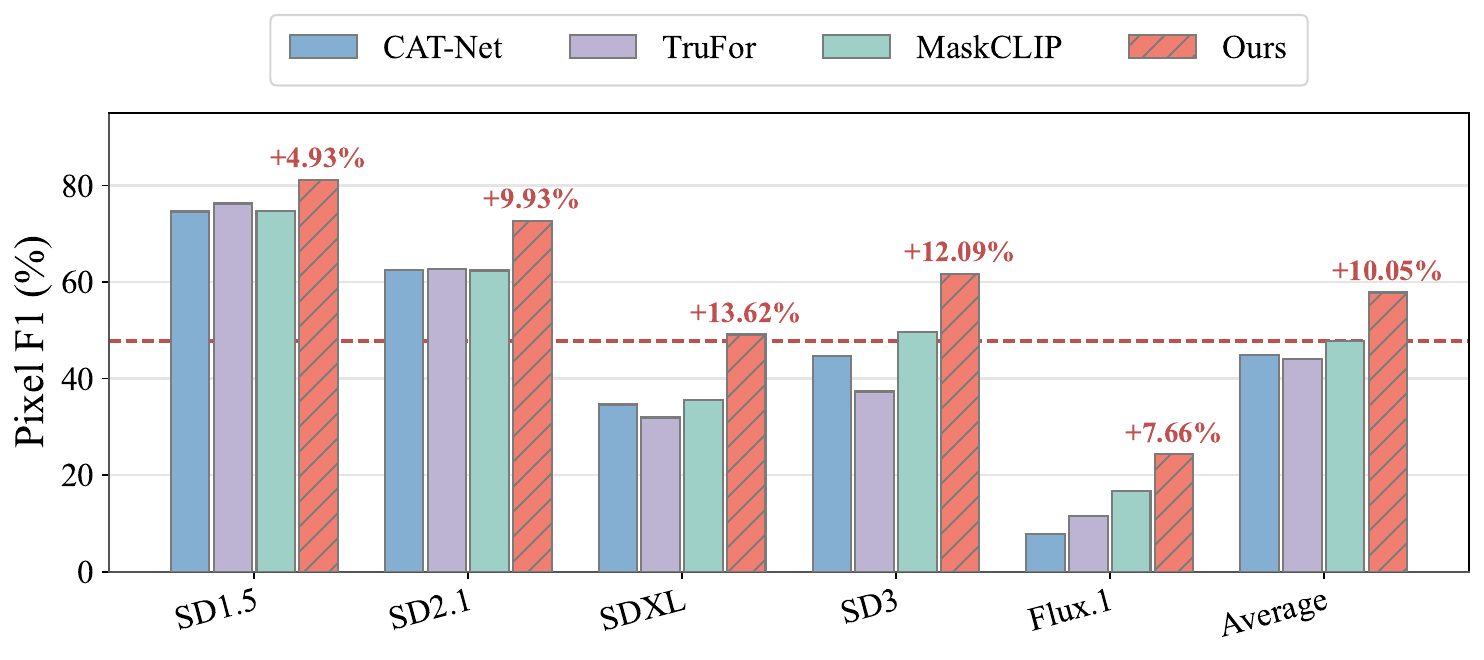}
  \Description{A bar chart comparing the Pixel F1 scores of various IML methods on the OpenSDI benchmark across different diffusion generators. The methods evaluated include traditional baselines such as CAT-Net and TruFor, as well as the diffusion-specific MaskCLIP. The bars for the proposed framework are noticeably taller across all generator categories, indicating superior localization performance compared to all other baseline methods.}
  \caption{\textbf{Localization performance on the OpenSDI benchmark measured by Pixel F1.} The proposed framework consistently outperforms previous traditional IML methods, including CAT-Net and TruFor, as well as the recent diffusion-oriented state-of-the-art method MaskCLIP across all evaluated generators.}
  \label{fig:performance_bar}
\end{figure}

As illustrated in Figure~\ref{fig:performance_bar}, existing methods exhibit architectural bias: artifact-driven models struggle with diffusion edits, while CLIP-based approaches lack explicit fine-grained frequency modeling. However, diffusion models still leave subtle spectral irregularities (e.g., via upsampling and latent decoding). Thus, frequency cues remain highly informative across both traditional statistical anomalies and generation-induced artifacts. Because current methods isolate microscopic traces from macroscopic semantic inconsistencies, a fundamental micro--macro gap emerges. Aligning these modalities remains a key challenge.

We propose \textbf{FASA} (\textbf{F}requency-\textbf{A}ware \textbf{S}emantic \textbf{A}lignment), a unified framework for localizing both traditional manipulations and diffusion-generated edits. Manipulation-sensitive frequency cues are extracted through an adaptive dual-band DCT decomposition, while multi-scale patch semantics are aggregated from a frozen CLIP~\cite{radford2021learning} image encoder. A patch-level contrastive alignment strategy is introduced to learn discriminative prototypes for authentic and manipulated regions in a shared latent space with text-guided initialization. A semantic-frequency side adapter injects multi-scale semantic priors into hierarchical frequency representations, and a frequency-gated mask decoder uses the optimized fake prototype as a semantic query to localize statistically suspicious and semantically inconsistent regions.

In summary, our main contributions are as follows:
\begin{itemize}
  \item We propose FASA, a unified framework that explicitly bridges microscopic frequency cues and macroscopic semantic reasoning for image manipulation localization across traditional manipulations and diffusion-generated edits.
  \item We introduce a dual-level alignment strategy that learns manipulation-aware semantic prototypes through contrastive alignment and integrates them with frequency evidence through prototype-guided decoding.
  \item Extensive experiments on both traditional manipulation and diffusion-generated benchmarks demonstrate state-of-the-art performance, strong generalization across settings, and robust localization under common degradations.
\end{itemize}

\section{Related Work}

\subsection{Image Manipulation Localization}
Early IML work focused on traditional manipulations leaving local forensic traces, with CNNs being the dominant approach. For example, MVSS-Net~\cite{chen2021image} uses boundary supervision, CAT-Net~\cite{kwon2022learning} exploits RGB and DCT for compression artifacts, PSCC-Net~\cite{liu2022pscc} performs progressive coarse-to-fine prediction, and APSC-Net~\cite{qu2024towards} combines adaptive perception and self-calibration.

Recent methods increasingly adopt Transformers to model long-range dependencies and global context. ObjectFormer~\cite{wang2022objectformer} combines RGB and high-frequency features with learnable object prototypes. IML-ViT~\cite{ma2023iml} establishes a pure ViT baseline for IML. TruFor~\cite{guillaro2023trufor} builds on learned noise-sensitive fingerprints and a transformer-style decoder. SparseViT~\cite{su2025can} promotes manipulation-sensitive non-semantic representations through sparse attention. Mesorch~\cite{zhu2025mesoscopic} integrates CNN and Transformer branches across multiple scales. DeCLIP~\cite{smeu2025declip} further shows that frozen CLIP representations can support dense deepfake localization through a learned decoder. Despite strong performance on conventional benchmarks, most of these methods are still tailored to traditional manipulations and often struggle with highly coherent diffusion-generated edits.

\subsection{Diffusion-Generated Forensics}
Visual forensics has increasingly shifted toward AI-generated content, predominantly focusing on image-level detection. CNNDet~\cite{wang2020cnn} and GramNet~\cite{liu2020global} use CNN representations and texture statistics, while FreqNet~\cite{tan2024frequency} emphasizes frequency artifacts. Recent methods improve cross-generator generalization via vision-language models; UniFD~\cite{ojha2023towards} uses CLIP with lightweight classifiers, and RINE~\cite{koutlis2024leveraging} exploits intermediate encoder-block features.

Pixel-level localization of diffusion-generated edits remains relatively underexplored. OpenSDI~\cite{wang2025opensdi} introduces an open-world benchmark for detecting and localizing such manipulations, and MaskCLIP~\cite{wang2025opensdi} combines CLIP and MAE~\cite{he2022masked} in this setting. Existing diffusion-oriented methods still emphasize either semantic reasoning or generator-specific artifacts, while explicit alignment between semantic priors and fine-grained frequency evidence remains limited. This gap motivates FASA, which unifies semantic alignment and frequency-aware localization across both traditional and diffusion-generated manipulations.

\begin{figure*}[t]
  \centering
  \includegraphics[width=\linewidth]{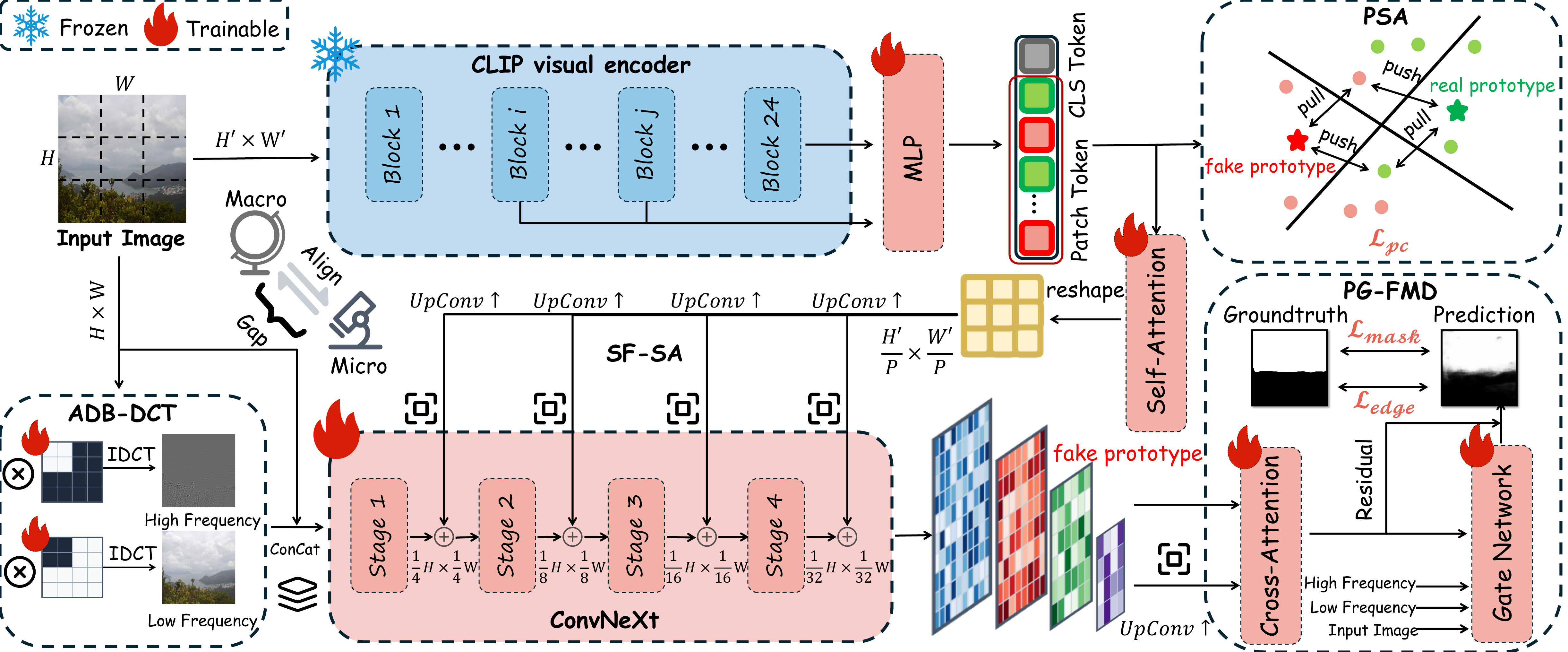}
  \Description{A detailed block diagram of the Frequency-Aware Semantic Alignment (FASA) framework. An input image is processed through two parallel branches. The top branch uses an Adaptive Dual-Band DCT (ADB-DCT) to extract high and low-frequency features. The bottom branch uses a frozen CLIP encoder and Patch-Level Semantic Alignment (PSA) to extract macroscopic semantic priors and optimize real and fake prototypes. A Semantic-Frequency Side Adapter (SF-SA) injects these semantic priors into the frequency branch at multiple scales. Finally, a Prototype-Guided Frequency-Gated Mask Decoder (PG-FMD) dynamically fuses the aligned features to output a predicted pixel-wise manipulation mask.}
    \caption{\textbf{Overview of FASA.} ADB-DCT extracts microscopic frequency cues, while a frozen CLIP encoder and PSA learn manipulation-aware semantic priors. SF-SA injects these priors into the hierarchical frequency pathway, and PG-FMD combines prototype-guided semantic responses with frequency gating for precise localization.}
  \label{fig:method}
\end{figure*}

\section{Proposed Method}

\subsection{Overview}
Manipulated regions, especially those generated by recent diffusion models, often appear visually plausible in the RGB domain while still leaving subtle spectral irregularities. Existing forensic pipelines frequently rely on fixed high-pass filters such as SRM or Laplacian operators. Although effective for exposing specific handcrafted traces, these filters are restricted to predefined frequency bands and may miss the diverse artifacts introduced by different editing pipelines and generative mechanisms. For this reason, an adaptive dual-band DCT module is introduced to learn manipulation-sensitive frequency cues in a data-driven manner.

As shown in Figure~\ref{fig:method}, the proposed framework follows a dual-level alignment paradigm. In the first stage, an Adaptive Dual-Band DCT module extracts manipulation-sensitive frequency cues, while the Patch-Level Semantic Alignment module derives manipulation-aware semantic priors from a frozen CLIP encoder. In the second stage, a Semantic-Frequency Side Adapter injects these priors into the hierarchical frequency pathway, and a Prototype-Guided Frequency-Gated Mask Decoder dynamically aggregates the aligned representations to predict the final mask. In this way, microscopic forensic evidence and macroscopic semantic reasoning are integrated within a unified localization framework.

\subsection{Frequency and Semantic Prior Construction}
The proposed framework is built on a clear division of roles between two complementary branches. The semantic branch provides manipulation-aware priors with strong global context, while the frequency branch preserves localized forensic sensitivity to subtle artifacts and boundary variations. Instead of merging these cues only at the final prediction stage, the two branches are connected progressively through prototype construction, multi-scale semantic injection, and prototype-guided decoding. This staged interaction allows semantic reasoning to guide localization without overwhelming the fine-grained evidence required for precise boundary delineation.

\textbf{Adaptive Dual-Band Frequency Decomposition (ADB-DCT).} Manipulated regions, especially from recent diffusion models, often appear visually plausible in RGB but leave subtle spectral irregularities. Existing pipelines frequently rely on fixed high-pass filters (e.g., SRM, Laplacian) that are restricted to predefined bands and may miss diverse artifacts introduced by varying generative mechanisms. Therefore, we introduce an adaptive dual-band DCT module to learn manipulation-sensitive frequency cues in a data-driven manner. The input image is first transformed into the DCT domain and decomposed into high-frequency and low-frequency components:
\begin{equation}
  \mathbf{I}^{h} = \mathrm{IDCT}(\mathbf{D} \odot \mathbf{M}_{h}(\alpha_h)), \quad
  \mathbf{I}^{l} = \mathrm{IDCT}(\mathbf{D} \odot \mathbf{M}_{l}(\alpha_l)),
\end{equation}
where $\mathbf{D}=\mathrm{DCT}(\mathbf{I})$, $\odot$ denotes element-wise multiplication, and $\mathbf{M}_{h}$ and $\mathbf{M}_{l}$ are sigmoid-based soft masks parameterized by learnable frequency cutoffs $\alpha_h$ and $\alpha_l$. This continuous formulation approximates hard frequency truncation while preserving differentiability and stable gradient propagation. The high-frequency branch emphasizes fine-grained forensic traces such as noise inconsistencies and generation artifacts, whereas the low-frequency branch preserves coarse structural layout and global content organization. The original RGB image is then concatenated with the two reconstructed frequency components to form a frequency-enhanced representation for the subsequent backbone.

\textbf{Patch-Level Semantic Alignment (PSA).} In parallel to the frequency branch, macroscopic semantic priors are extracted using a frozen CLIP image encoder. Instead of relying only on the final-layer representation, patch tokens are aggregated from selected intermediate and deep layers so that both local structural cues and high-level semantic information can be retained. Let $\mathbf{t}_i^{l}$ denote the patch token at spatial location $i$ from the $l$-th selected CLIP layer. The tokens from these layers are concatenated and projected into a shared embedding space:
\begin{equation}
  \mathbf{p}_i = \mathbf{W}_{m} \Big( \mathop{\big\|}_{l \in \mathcal{S}} \mathbf{t}_i^{l} \Big),
\end{equation}
where $\mathcal{S}$ denotes the selected layer set and $\|$ denotes channel-wise concatenation. To strengthen contextual interaction among patches, the projected token sequence is further refined by a multi-head self-attention module:
\begin{equation}
  \hat{\mathbf{P}} = \mathrm{MSA}(\mathbf{P}) + \mathbf{P}.
\end{equation}
The refined tokens are then rearranged into a dense semantic feature map $\mathbf{F}_{c}$ that captures both local visual structure and global semantic coherence. To make this representation explicitly aware of manipulation, a real prototype $\mathbf{e}_{r}$ and a fake prototype $\mathbf{e}_{f}$ are initialized in the shared latent space using CLIP text embeddings generated from prompts such as ``an authentic image region'' and ``a manipulated image region'', and are subsequently optimized during training as task-specific semantic anchors for authentic and manipulated regions. A linear projector then maps each patch feature in $\mathbf{F}_{c}$ to a visual embedding $\mathbf{z}_i$, and patch-to-prototype alignment is enforced with the following InfoNCE objective:
\begin{equation}
  \mathcal{L}_{pc} = -\frac{1}{N}\sum_{i=1}^{N}
  \log
  \frac{\exp(\mathrm{sim}(\mathbf{z}_i,\mathbf{e}_{y_i})/\tau)}
  {\sum_{c\in\{r,f\}} \exp(\mathrm{sim}(\mathbf{z}_i,\mathbf{e}_{c})/\tau)},
\end{equation}
where $y_i$ is the supervision label of the $i$-th patch, $\mathrm{sim}(\cdot,\cdot)$ denotes cosine similarity, and $\tau$ is a learnable temperature parameter. This alignment encourages authentic and manipulated regions to occupy distinct semantic subspaces, improves feature separability under both traditional and diffusion-generated manipulations, and produces a task-adapted fake prototype for subsequent mask decoding. From the perspective of representation learning, PSA serves two roles simultaneously: it strengthens the discriminability of patch-level semantic features and also supplies an explicit semantic query for dense localization. This coupling between semantic supervision and mask prediction is important for maintaining consistency between semantic alignment and final localization.

\subsection{Prototype-Guided Mask Decoding}

\textbf{Semantic-Frequency Side Adapter (SF-SA).} The frequency-enhanced representation is fed into a hierarchical convolutional side backbone whose first stem is expanded to accommodate the augmented input channels. This side pathway produces multi-scale frequency features at progressively reduced resolutions, but these features mainly encode local forensic evidence and do not directly model semantic inconsistency. To explicitly connect the two modalities, the semantic map $\mathbf{F}_{c}$ is injected into each stage of the frequency pathway through a semantic-frequency side adapter:
\begin{equation}
  \tilde{\mathbf{F}}_{s}^{k} = \mathbf{F}_{s}^{k} + \mathcal{A}_{k}(\mathbf{F}_{c}), \quad k \in \{1,2,3,4\},
\end{equation}
where $\mathbf{F}_{s}^{k}$ denotes the frequency feature at stage $k$, and $\mathcal{A}_{k}(\cdot)$ denotes the adaptation module. Specifically, the adapter progressively upsamples the semantic map to the target spatial resolution using transposed convolutions, and then applies a $1\times1$ convolution followed by a $3\times3$ convolution to align channel capacity and smooth the adapted representation. Layer normalization is inserted between these operations to stabilize optimization. This design allows semantic priors to be injected into the hierarchical frequency pathway in a structurally compatible manner, so that low-level forensic sensitivity is preserved while high-level semantic guidance is introduced at multiple scales.

\begin{table*}[t]
  \caption{Quantitative comparison of pixel-level localization performance on the OpenSDI benchmark. Pixel F1 and IoU are reported for each diffusion generator, and Average denotes the mean performance across all generators. Bold indicates the best result, and underline indicates the second-best result.}
  \label{tab:diffusion-loc}
  \centering
  \begin{tabular}{lcccccccccccc}
    \toprule
    \multirow{2}{*}{Method} & \multicolumn{2}{c}{SD1.5} & \multicolumn{2}{c}{SD2.1} & \multicolumn{2}{c}{SDXL} & \multicolumn{2}{c}{SD3} & \multicolumn{2}{c}{Flux.1} & \multicolumn{2}{c}{Average} \\
    \cmidrule(lr){2-3} \cmidrule(lr){4-5} \cmidrule(lr){6-7} \cmidrule(lr){8-9} \cmidrule(lr){10-11} \cmidrule(lr){12-13}
    & F1 & IoU & F1 & IoU & F1 & IoU & F1 & IoU & F1 & IoU & F1 & IoU \\
    \midrule
    MVSS-Net~\cite{chen2021image} & 0.6420 & 0.5656 & 0.5011 & 0.4299 & 0.2202 & 0.1758 & 0.3569 & 0.2950 & 0.0889 & 0.0674 & 0.3618 & 0.3067 \\
    PSCC-Net~\cite{liu2022pscc} & 0.6243 & 0.5156 & 0.5029 & 0.4055 & 0.3323 & 0.2514 & 0.4135 & 0.3215 & 0.1238 & 0.0850 & 0.3994 & 0.3158 \\
    CAT-Net~\cite{kwon2022learning} & 0.7460 & 0.6589 & 0.6254 & 0.5450 & 0.3464 & 0.2861 & 0.4459 & 0.3764 & 0.0777 & 0.0583 & 0.4483 & 0.3849 \\
    TruFor~\cite{guillaro2023trufor} & 0.7626 & 0.6862 & 0.6278 & 0.5540 & 0.3195 & 0.2636 & 0.3733 & 0.3101 & 0.1160 & 0.0891 & 0.4398 & 0.3806 \\
    APSC-Net~\cite{qu2024towards} & 0.7556 & 0.6909 & 0.6110 & 0.5464 & 0.3287 & 0.2809 & 0.4256 & 0.3690 & 0.1181 & 0.0949 & 0.4478 & 0.3964 \\
    IML-ViT~\cite{ma2023iml} & 0.7724 & 0.7031 & 0.5940 & 0.5266 & 0.3151 & 0.2626 & 0.3502 & 0.2933 & 0.0910 & 0.0707 & 0.4245 & 0.3713 \\
    Mesorch~\cite{zhu2025mesoscopic} & \underline{0.7998} & \underline{0.7319} & \underline{0.6344} & \underline{0.5709} & 0.3232 & 0.2781 & 0.3735 & 0.3238 & 0.0863 & 0.0696 & 0.4434 & 0.3949 \\
    SparseViT~\cite{su2025can} & 0.7539 & 0.6854 & 0.5845 & 0.5207 & 0.3396 & 0.2920 & 0.4020 & 0.3485 & 0.0987 & 0.0801 & 0.4357 & 0.3853 \\
    MaskCLIP~\cite{wang2025opensdi} & 0.7463 & 0.6731 & 0.6239 & 0.5530 & \underline{0.3552} & \underline{0.2996} & \underline{0.4965} & \underline{0.4254} & \underline{0.1671} & \underline{0.1330} & \underline{0.4778} & \underline{0.4168} \\
    \rowcolor{gray!10} Ours & \textbf{0.8119} & \textbf{0.7340} & \textbf{0.7271} & \textbf{0.6431} & \textbf{0.4914} & \textbf{0.4103} & \textbf{0.6174} & \textbf{0.5285} & \textbf{0.2437} & \textbf{0.1885} & \textbf{0.5783} & \textbf{0.5009} \\
    \bottomrule
  \end{tabular}
\end{table*}

\textbf{Prototype-Guided Frequency-Gated Mask Decoder (PG-FMD).} Given the aligned multi-scale features $\{\tilde{\mathbf{F}}_{s}^{k}\}_{k=1}^{4}$, the decoder first upsamples them to a common resolution and then performs cross-attention between the semantic prototype and the spatial feature maps. For each scale, the fake prototype is linearly projected as the semantic query, while the aligned feature map provides the keys and values:
\begin{equation}
  \mathbf{B}_{k} = \mathrm{Proj}\Big(\mathrm{CrossAttn}(\mathbf{Q}_k,\mathbf{K}_k,\mathbf{V}_k)\Big).
\end{equation}
The resulting basis mask $\mathbf{B}_{k}$ reflects the response of semantically suspicious regions at each scale. However, semantic queries alone may lead to over-segmentation, especially when visually salient but authentic regions trigger strong semantic responses. To alleviate this issue, a spatial gating network is introduced to operate on the frequency-enhanced input and predict scale-wise weights for these basis masks. The final pre-activation logit map is computed as
\begin{equation}
  \mathbf{O} = \sum_{k=1}^{4} \mathbf{G}_{k} \odot \mathbf{B}_{k} + \phi(\mathbf{B}),
\end{equation}
where $\mathbf{G}$ denotes the predicted gating map, $\mathbf{B}$ is the concatenation of all basis masks, and $\phi(\cdot)$ is a $1\times1$ convolution used to preserve holistic semantic evidence. This formulation contains two complementary pathways: the gated summation term uses frequency evidence to refine local structure and sharpen manipulation boundaries, while the residual semantic pathway preserves global manipulation cues and serves as a fallback when forensic traces are weakened by degradations such as JPEG compression or blur. The final manipulation mask is obtained by upsampling $\mathbf{O}$ to the original image resolution and applying a sigmoid function:
\begin{equation}
  \hat{\mathbf{M}} = \sigma(\mathrm{Up}(\mathbf{O})).
\end{equation}
This decoder design is well aligned with the visual characteristics of the two manipulation paradigms. For diffusion-generated edits, semantic inconsistency often provides stronger evidence than obvious local artifacts, so the prototype-guided query helps highlight suspicious content regions. For traditional manipulations, low-level traces and boundary discontinuities remain highly informative, so frequency gating plays an important role in suppressing false positives and refining mask contours. By combining these effects within the same decoder, the proposed framework remains effective across both semantically deceptive diffusion edits and artifact-rich traditional manipulations.

Based on the predicted manipulation mask, the framework also performs image-level detection so that localization and classification remain tightly coupled. Following standard evaluation practice, the image-level score is defined as the maximum predicted probability over all spatial locations:
\begin{equation}
  S = \max_{h,w} \hat{\mathbf{M}}(h,w).
\end{equation}
The input image is classified as manipulated if $S$ exceeds a predefined threshold. This unified formulation preserves consistency between pixel-level localization and image-level classification.

\subsection{Overall Optimization Objective}
The proposed framework is optimized end-to-end with a multi-term objective. Following prior IML works~\cite{ma2023iml, wang2025opensdi}, dense localization is supervised by a binary cross-entropy loss on the predicted mask and an edge-aware loss that emphasizes manipulation boundaries. Together with the prototype contrastive loss, the overall objective is written as
\begin{equation}
  \mathcal{L} = \lambda_{m}\mathcal{L}_{mask} + \lambda_{e}\mathcal{L}_{edge} + \lambda_{pc}\mathcal{L}_{pc},
\end{equation}
where $\lambda_{m}$, $\lambda_{e}$, and $\lambda_{pc}$ are balancing coefficients. These three terms play complementary roles: $\mathcal{L}_{mask}$ supervises dense region prediction, $\mathcal{L}_{edge}$ improves boundary precision, and $\mathcal{L}_{pc}$ enforces manipulation-aware semantic alignment. Their joint optimization enables the proposed framework to learn semantically discriminative, frequency-sensitive, and spatially precise representations for robust image manipulation localization.

\begin{table*}[t]
  \caption{Quantitative comparison of image-level detection performance on the OpenSDI benchmark. Image F1 and Accuracy are reported for each diffusion generator, and Average denotes the mean performance across all generators. The upper group contains image-level detection methods, and the lower group contains pixel-level localization methods. Bold indicates the best result, and underline indicates the second-best result.}
  \label{tab:diffusion-det}
  \centering
  \begin{tabular}{lcccccccccccc}
    \toprule
    \multirow{2}{*}{Method} & \multicolumn{2}{c}{SD1.5} & \multicolumn{2}{c}{SD2.1} & \multicolumn{2}{c}{SDXL} & \multicolumn{2}{c}{SD3} & \multicolumn{2}{c}{Flux.1} & \multicolumn{2}{c}{Average} \\
    \cmidrule(lr){2-3} \cmidrule(lr){4-5} \cmidrule(lr){6-7} \cmidrule(lr){8-9} \cmidrule(lr){10-11} \cmidrule(lr){12-13}
    & F1 & Acc & F1 & Acc & F1 & Acc & F1 & Acc & F1 & Acc & F1 & Acc \\
    \midrule
    CNNDet~\cite{wang2020cnn} & 0.8460 & 0.8504 & 0.7156 & 0.7594 & 0.5970 & 0.6872 & 0.5627 & 0.6708 & 0.3572 & 0.5757 & 0.6157 & 0.7087 \\
    GramNet~\cite{liu2020global} & 0.8051 & 0.8035 & 0.7401 & 0.7666 & 0.6528 & 0.7076 & 0.6435 & 0.7029 & 0.5200 & 0.6337 & 0.6723 & 0.7229 \\
    UniFD~\cite{ojha2023towards} & 0.7745 & 0.7760 & 0.8062 & 0.8192 & 0.7074 & 0.7483 & 0.7109 & 0.7517 & 0.6110 & 0.6906 & 0.7220 & 0.7572 \\
    FreqNet~\cite{tan2024frequency} & 0.7588 & 0.7770 & 0.6097 & 0.6837 & 0.5315 & 0.6402 & 0.5350 & 0.6437 & 0.3847 & 0.5708 & 0.5639 & 0.6631 \\
    NPR~\cite{tan2024rethinking} & 0.7941 & 0.7928 & 0.8167 & 0.8184 & 0.7212 & 0.7428 & 0.7343 & 0.7547 & \textbf{0.6762} & \textbf{0.7136} & 0.7485 & 0.7645 \\
    RINE~\cite{koutlis2024leveraging} & 0.9108 & 0.9098 & 0.8747 & 0.8812 & 0.7343 & 0.7876 & 0.7205 & 0.7678 & 0.5586 & 0.6702 & \underline{0.7598} & 0.8033 \\
    \midrule
    MVSS-Net~\cite{chen2021image} & 0.9194 & 0.9205 & 0.7674 & 0.7998 & 0.5886 & 0.6912 & 0.6702 & 0.7388 & 0.3788 & 0.5976 & 0.6649 & 0.7496 \\
    PSCC-Net~\cite{liu2022pscc} & 0.9576 & 0.9579 & 0.7773 & 0.8138 & 0.6217 & 0.7207 & 0.6040 & 0.7116 & 0.4584 & 0.6426 & 0.6838 & 0.7693 \\
    CAT-Net~\cite{kwon2022learning} & 0.9494 & 0.9481 & 0.8378 & 0.8531 & \underline{0.7449} & 0.7868 & \underline{0.7509} & 0.7911 & 0.3583 & 0.5897 & 0.7283 & 0.7938 \\
    TruFor~\cite{guillaro2023trufor} & 0.9555 & 0.9545 & \underline{0.8817} & \underline{0.8891} & 0.6556 & 0.7314 & 0.6845 & 0.7491 & 0.3505 & 0.5877 & 0.7056 & 0.7824 \\
    APSC-Net~\cite{qu2024towards} & 0.9549 & 0.9557 & 0.8239 & 0.8465 & 0.4996 & 0.6579 & 0.5995 & 0.7088 & 0.2953 & 0.5771 & 0.6346 & 0.7492 \\
    IML-ViT~\cite{ma2023iml} & \underline{0.9645} & \underline{0.9642} & 0.8668 & 0.8788 & 0.7291 & 0.7802 & 0.7459 & \underline{0.7923} & 0.4581 & 0.6384 & 0.7529 & \underline{0.8108} \\
    Mesorch~\cite{zhu2025mesoscopic} & \textbf{0.9729} & \textbf{0.9732} & 0.7942 & 0.8279 & 0.5541 & 0.6885 & 0.5555 & 0.6897 & 0.2446 & 0.5665 & 0.6242 & 0.7492 \\
    SparseViT~\cite{su2025can} & 0.9597 & 0.9604 & 0.7865 & 0.8214 & 0.5707 & 0.6955 & 0.6383 & 0.7309 & 0.2983 & 0.5822 & 0.6507 & 0.7581 \\
    MaskCLIP~\cite{wang2025opensdi} & 0.9127 & 0.9143 & 0.8595 & 0.8728 & 0.7431 & \underline{0.7882} & 0.7007 & 0.7634 & 0.5484 & 0.6798 & 0.7529 & 0.8037 \\
    \rowcolor{gray!10} Ours & 0.9375 & 0.9343 & \textbf{0.9391} & \textbf{0.9373} & \textbf{0.8926} & \textbf{0.8931} & \textbf{0.8774} & \textbf{0.8810} & \underline{0.6282} & \underline{0.7041} & \textbf{0.8550} & \textbf{0.8700} \\
    \bottomrule
  \end{tabular}
\end{table*}

\begin{table*}[t]
  \caption{Quantitative comparison of pixel-level localization performance on four traditional manipulation benchmarks. Pixel F1 and AUC are reported for each dataset, and Average denotes the mean performance across all evaluated datasets. Bold indicates the best result, and underline indicates the second-best result.}
  \label{tab:traditional-loc}
  \centering
  \begin{tabular}{lcccccccccc}
    \toprule
    \multirow{2}{*}{Method} & \multicolumn{2}{c}{CASIAv1} & \multicolumn{2}{c}{Columbia} & \multicolumn{2}{c}{Coverage} & \multicolumn{2}{c}{NIST16} & \multicolumn{2}{c}{Average} \\
    \cmidrule(lr){2-3} \cmidrule(lr){4-5} \cmidrule(lr){6-7} \cmidrule(lr){8-9} \cmidrule(lr){10-11}
    & F1 & AUC & F1 & AUC & F1 & AUC & F1 & AUC & F1 & AUC \\
    \midrule
    MVSS-Net~\cite{chen2021image} & 0.5775 & 0.9117 & 0.7274 & 0.9237 & 0.4663 & 0.8632 & 0.3301 & 0.7883 & 0.5253 & 0.8717 \\
    PSCC-Net~\cite{liu2022pscc} & 0.5690 & 0.9188 & 0.8555 & 0.9458 & 0.4034 & 0.8838 & 0.3655 & 0.8279 & 0.5484 & 0.8941 \\
    CAT-Net~\cite{kwon2022learning} & 0.8081 & 0.9804 & 0.9150 & 0.9457 & 0.4273 & 0.9168 & 0.2522 & 0.8216 & 0.6007 & 0.9161 \\
    TruFor~\cite{guillaro2023trufor} & 0.8208 & 0.9742 & 0.8754 & 0.8996 & 0.4506 & \underline{0.9424} & 0.3359 & 0.8782 & 0.6207 & 0.9236 \\
    APSC-Net~\cite{qu2024towards} & 0.8494 & 0.9772 & \underline{0.9299} & 0.9558 & 0.4539 & 0.9284 & 0.2597 & 0.8535 & 0.6232 & 0.9287 \\
    IML-ViT~\cite{ma2023iml} & 0.8137 & 0.9726 & 0.8864 & 0.9544 & \underline{0.6135} & 0.9407 & 0.3983 & 0.8828 & 0.6780 & \underline{0.9376} \\
    Mesorch~\cite{zhu2025mesoscopic} & 0.8398 & 0.9852 & 0.8905 & 0.9095 & 0.5862 & 0.9327 & 0.3921 & \underline{0.8883} & 0.6772 & 0.9289 \\
    SparseViT~\cite{su2025can} & 0.8196 & 0.9687 & \textbf{0.9568} & \textbf{0.9770} & 0.5165 & 0.9386 & 0.3726 & 0.8508 & 0.6664 & 0.9338 \\
    MaskCLIP~\cite{wang2025opensdi} & \underline{0.8627} & \underline{0.9858} & 0.8681 & 0.8981 & 0.5691 & 0.9382 & \textbf{0.4345} & \textbf{0.8884} & \underline{0.6836} & 0.9276 \\
    \rowcolor{gray!10} Ours & \textbf{0.8777} & \textbf{0.9900} & 0.9057 & \underline{0.9577} & \textbf{0.6537} & \textbf{0.9562} & \underline{0.4092} & 0.8768 & \textbf{0.7116} & \textbf{0.9452} \\
    \bottomrule
  \end{tabular}
\end{table*}

\begin{figure*}[t]
  \centering
  \includegraphics[width=\linewidth]{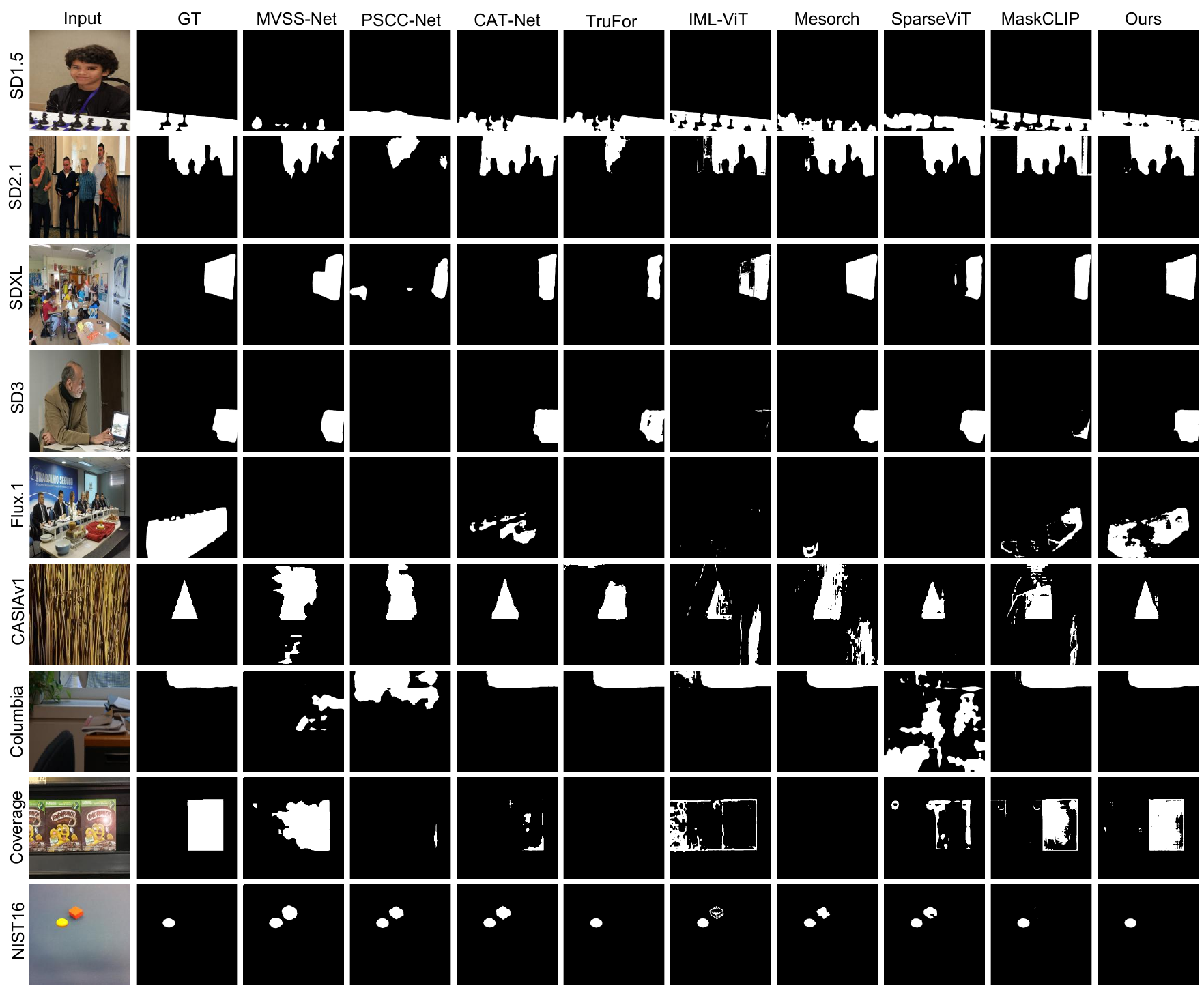}
  \Description{A grid of images demonstrating qualitative localization results. The rows show different image samples from diffusion-generated and traditional manipulation datasets. The columns display the original image, ground truth mask, and predicted masks from various baseline methods and the proposed framework. The baselines frequently exhibit severe missed detections (showing blank masks) or over-segmentation (showing noisy false positives), whereas the proposed framework consistently produces precise and complete localization masks that closely align with the ground truth.}
  \caption{Qualitative comparison of localization results on diffusion-generated and traditional manipulation datasets. Existing methods often suffer from missed detections or over-segmentation, whereas the proposed framework produces more complete masks and more precise boundaries across diverse manipulation types.}
  \label{fig:qualitative}
\end{figure*}

\begin{figure*}[t]
  \centering
  \includegraphics[width=\textwidth]{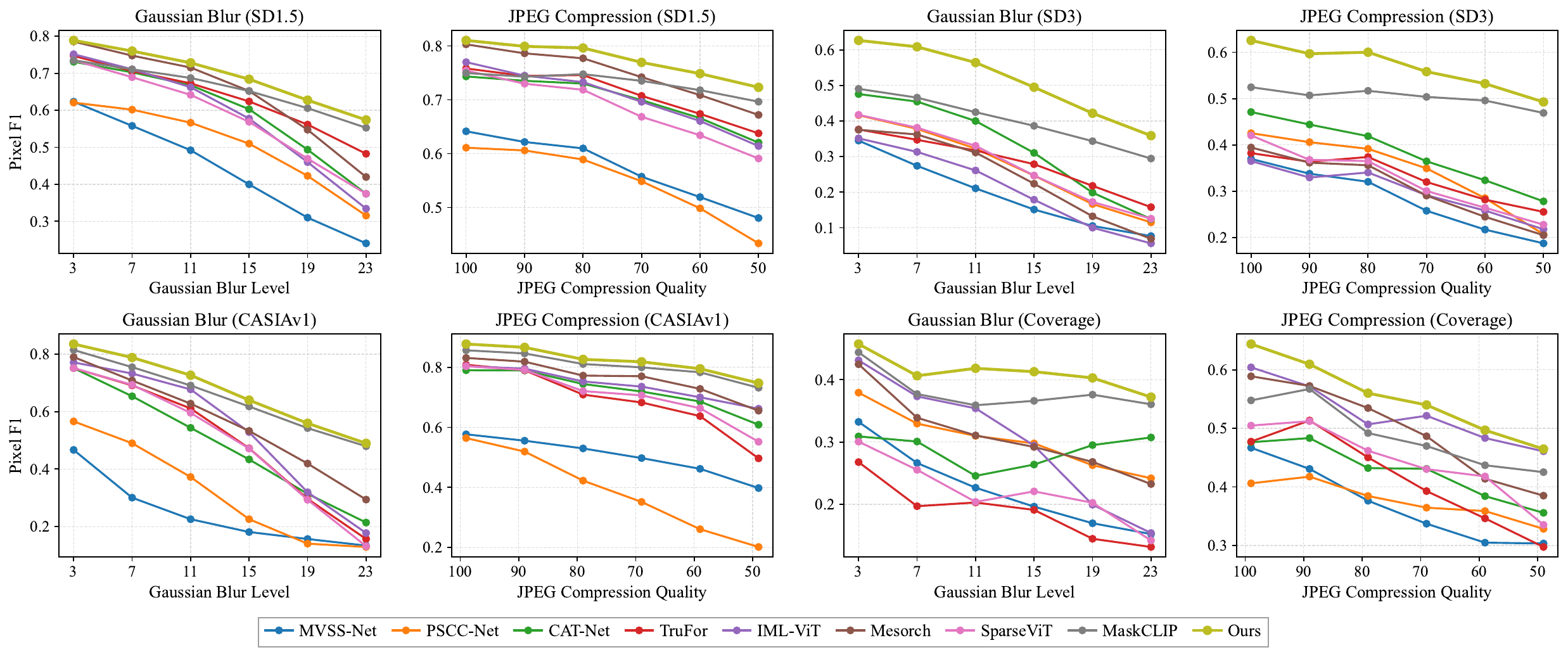}
  \Description{Four line graphs evaluating robustness against Gaussian blur and JPEG compression at varying intensity levels. The x-axis represents degradation severity, and the y-axis represents the Pixel-level  F1 score. The top row shows results on the OpenSDI dataset (SD1.5 and SD3), and the bottom row shows results on traditional datasets (CASIAv1 and Coverage). In all four graphs, the curve representing the proposed framework starts higher and decays at a significantly slower rate than the curves of competing baseline models as the degradation severity increases.}
  \caption{Robustness comparison of different methods under Gaussian blur and JPEG compression. The first row reports results on OpenSDI with SD1.5 and SD3, while the second row reports results on the IML benchmark with CASIAv1 and Coverage. Pixel-level F1 is used for evaluation, and higher values indicate better robustness.}
  \label{fig:robustness_all}
\end{figure*}

\section{Experiments}

\subsection{Experimental Setup}

\textbf{Datasets.}
Experiments are conducted on two complementary settings. For diffusion-generated edits, we use the OpenSDI benchmark~\cite{wang2025opensdi}. The model is trained on authentic images and SD1.5-based edited samples. Cross-generator generalization is then evaluated on test sets built from authentic images and manipulations produced by five generators, namely SD1.5~\cite{rombach2022high}, SD2.1~\cite{rombach2022high}, SDXL~\cite{podell2024sdxl}, SD3~\cite{esser2024scaling}, and Flux.1~\cite{flux2024}. For traditional manipulations, training follows the CAT-Net protocol~\cite{kwon2022learning} adopted by recent state-of-the-art methods~\cite{su2025can, zhu2025mesoscopic}. The training set combines CASIAv2~\cite{dong2013casia}, FantasticReality~\cite{kniaz2019point}, IMD2020~\cite{novozamsky2020imd2020}, and TampCOCO~\cite{kwon2022learning}. Cross-dataset evaluation is conducted on CASIAv1~\cite{dong2013casia}, Columbia~\cite{ng2009columbia}, Coverage~\cite{wen2016coverage}, and NIST16~\cite{guan2019mfc} without fine-tuning.

\textbf{Evaluation Metrics.}
For diffusion-generated edits, we follow the evaluation protocol of OpenSDI. Pixel-level localization is evaluated with pixel-level F1 and IoU after thresholding the predicted masks at 0.5, while image-level detection is evaluated with image-level F1 and Accuracy. For traditional manipulations, we follow the common IMDL-BenCo protocol and report pixel-level F1 and AUC to evaluate cross-dataset localization performance.

\textbf{Baseline Methods.}
The proposed framework is compared with 15 state-of-the-art methods. For pixel-level localization on both settings, nine representative models are considered: MVSS-Net~\cite{chen2021image}, PSCC-Net~\cite{liu2022pscc}, CAT-Net~\cite{kwon2022learning}, TruFor~\cite{guillaro2023trufor}, APSC-Net~\cite{qu2024towards}, IML-ViT~\cite{ma2023iml}, Mesorch~\cite{zhu2025mesoscopic}, SparseViT~\cite{su2025can}, and MaskCLIP~\cite{wang2025opensdi}. For image-level detection, six dedicated binary detectors are additionally included: CNNDet~\cite{wang2020cnn}, GramNet~\cite{liu2020global}, UniFD~\cite{ojha2023towards}, FreqNet~\cite{tan2024frequency}, NPR~\cite{tan2024rethinking}, and RINE~\cite{koutlis2024leveraging}. To ensure fair comparison, all localization baselines are retrained and evaluated under the same setting on OpenSDI, while all localization methods on traditional manipulation benchmarks are trained and evaluated within the unified IMDL-BenCo framework~\cite{ma2024imdl}.

\textbf{Implementation Details.}
Macroscopic semantic priors are extracted from selected intermediate and deep layers of a frozen CLIP-ViT-L/14~\cite{radford2021learning} encoder, which provide complementary texture, structural, and semantic information, while the microscopic frequency pathway is built on ConvNeXt-Base~\cite{liu2022convnet}. During training, images are augmented with Gaussian blur, JPEG compression, random scaling, random cropping, and horizontal flipping. All images are resized to $224\times224$ for CLIP and $512\times512$ for ConvNeXt. The whole framework is trained end-to-end for 20 epochs with AdamW, using a weight decay of 0.05 and a batch size of 32. The initial learning rate is set to $1\times10^{-4}$ and decayed with a cosine annealing schedule. The loss weights are set to $\lambda_m=1.0$, $\lambda_{pc}=1.0$, and $\lambda_e=20.0$. The image-level decision threshold is set to 0.5. All experiments are conducted on four RTX 4090 GPUs.

\subsection{Performance Comparisons}

\textbf{Pixel-Level Localization on Diffusion-Generated Edits.}
Table~\ref{tab:diffusion-loc} reports localization results on OpenSDI, which provides a rigorous test of cross-generator generalization. Although many existing methods perform competitively on SD1.5, their accuracy drops substantially on unseen generators. In contrast, the proposed framework establishes state-of-the-art performance across all generators, achieving an average pixel-level F1 of 0.5783 and IoU of 0.5009. These scores outperform the second-best method by 10.05\% in F1 and 8.41\% in IoU. The benefit becomes clear on the more challenging cross-generator settings; even on Flux.1, the most difficult setting in the benchmark, the proposed framework achieves the best F1 score of 0.2437, exceeding the second-best result by 7.66\%. These results indicate that explicit alignment between semantic priors and microscopic frequency evidence is highly effective for exposing subtle traces left by diverse diffusion generators.

\textbf{Image-Level Detection on Diffusion-Generated Edits.}
Table~\ref{tab:diffusion-det} presents the image-level detection results on OpenSDI. A similar generalization pattern is observed in image-level detection. Traditional IML models tend to overfit the SD1.5 training domain and suffer from clear generalization drops on newer generators, while dedicated binary detectors provide only limited overall robustness. The proposed framework outperforms all detection baselines and localization-based competitors, achieving an average F1 of 0.8550 and an average Accuracy of 0.8700. These scores exceed the second-best results by 9.52\% in F1 and 5.92\% in Accuracy. Although the proposed framework does not achieve its largest advantage on the SD1.5 intra-generator setting, the benefit becomes clear on cross-generator evaluation. This result suggests that the semantic alignment module improves not only dense localization but also the aggregation of manipulation evidence at the image level.

\textbf{Pixel-Level Localization on Traditional Manipulations.}
Table~\ref{tab:traditional-loc} shows the cross-dataset localization results on traditional manipulation benchmarks. Strong performance is maintained in this setting as well. The proposed framework achieves the best average F1 of 0.7116 and the best average AUC of 0.9452, surpassing previous state-of-the-art methods overall. The advantage is particularly clear on the challenging Coverage and CASIAv1 benchmarks, where the improvement over the corresponding second-best F1 result reaches 4.02\% and 1.50\%, respectively. Competitive performance is also observed on Columbia and NIST16. These results indicate that the proposed framework does not sacrifice performance on conventional manipulations while extending effectively to diffusion-generated edits.

\textbf{Qualitative Comparison.}
As illustrated in Figure~\ref{fig:qualitative}, existing methods often struggle to balance sensitivity to manipulated regions with suppression of false positives across different manipulation types. On advanced diffusion-generated edits such as SD3 and Flux.1, artifact-driven baselines frequently miss manipulated content because the synthetic regions remain highly coherent in local texture and statistics. On traditional manipulation datasets with cluttered backgrounds, recent state-of-the-art methods including IML-ViT, SparseViT, and Mesorch often produce severe over-segmentation. In contrast, the proposed framework yields masks that are both more complete and more precise. It reduces missed detections in semantically suspicious yet visually smooth diffusion-generated regions, while suppressing background activation and preserving compact foreground masks on traditional benchmarks. These observations are consistent with the design intuition of the proposed framework, where semantic priors improve sensitivity and frequency evidence controls spatial precision. This advantage is particularly visible in challenging cases where manipulated regions are either smoothly blended into the surrounding content or embedded in visually complex backgrounds.

\subsection{Robustness Evaluation}
Robustness is further evaluated under two common image degradations, namely Gaussian blur and JPEG compression at different severity levels. Figure~\ref{fig:robustness_all} reports the corresponding pixel-level F1 scores on both diffusion-generated edits and traditional manipulations. On diffusion-generated data, the proposed framework maintains a clear advantage under all degradation settings, with the margin being especially pronounced in the SD3 cross-generator case as competing methods degrade rapidly. A similar pattern is observed on traditional manipulation benchmarks, particularly under JPEG compression on CASIAv1 and Gaussian blur on Coverage. As image quality deteriorates, the gap between the proposed framework and artifact-driven competitors becomes increasingly large. These results show that semantic-frequency alignment improves robustness across both diffusion-generated and traditional manipulations: semantic priors remain stable when forensic traces are corrupted, while the frequency pathway still provides useful local evidence when image quality is less severely degraded.

\subsection{Ablation Study}
Table~\ref{tab:ablation-study} reports a progressive ablation study on OpenSDI. Starting from a pure spatial baseline composed of a ConvNeXt backbone and a simple decoder, the average pixel-level F1 and IoU are 0.4303 and 0.3758, respectively. Introducing ADB-DCT improves the scores to 0.4607 and 0.4039, showing that adaptive frequency decomposition provides useful microscopic evidence for localization. Adding the frozen CLIP encoder together with PSA further raises the performance to 0.4917 in F1 and 0.4289 in IoU, indicating that manipulation-aware semantic priors improve feature discriminability and semantic consistency. After SF-SA is incorporated, the performance increases to 0.5128 in F1 and 0.4388 in IoU, confirming the benefit of injecting semantic priors into the frequency pathway at multiple scales. Replacing the simple decoder with PG-FMD yields the best performance of 0.5783 in F1 and 0.5009 in IoU. The consistent gains across all stages validate the necessity of each component in the full architecture.

\begin{table}[t]
  \caption{Ablation study of main components in FASA on the OpenSDI benchmark (average performance).}
  \label{tab:ablation-study}
  \centering
  \begin{tabular}{lcc}
    \toprule
    Model Variation & Pixel F1 & Pixel IoU \\
    \midrule
    Baseline & 0.4303 & 0.3758 \\
    + ADB-DCT & 0.4607 & 0.4039 \\
    + PSA & 0.4917 & 0.4289 \\
    + SF-SA & 0.5128 & 0.4388 \\
    + PG-FMD (Full FASA) & 0.5783 & 0.5009 \\
    \bottomrule
  \end{tabular}
\end{table}

\section{Conclusions}
This paper presents FASA, a unified framework for image manipulation localization across both traditional manipulations and diffusion-generated edits. By combining manipulation-aware semantic alignment with adaptive frequency modeling, the proposed framework explicitly bridges the micro--macro gap between macroscopic semantic inconsistency and microscopic forensic evidence. Extensive experiments on multiple benchmarks demonstrate state-of-the-art localization performance, strong cross-generator and cross-dataset generalization, and robust behavior under common image degradations like blur and compression. These findings highlight the importance of jointly exploiting semantic priors and frequency cues for robust localization across diverse and increasingly challenging manipulation paradigms.

\begin{acks}
This work is supported by the National Natural Science Foundation of China (No.62441237).
\end{acks}

\bibliographystyle{ACM-Reference-Format}
\bibliography{main}

\end{document}